# Predicting Power Electronics Device Reliability under Extreme Conditions with Machine Learning Algorithms

Carlos A. Olivares, Raziur Rahman, Christopher Stankus, Jade Hampton, Andrew Zedwick, Moinuddin Ahmed

*Abstract*— Power device reliability is a major concern during operation under extreme environments, as doing so reduces the operational lifetime of any power system or sensing infrastructure. Due to a potential for system failure, devices must be experimentally validated before implementation, which is expensive and time-consuming. In this paper, we have utilized machine learning algorithms to predict device reliability, significantly reducing the need for conducting experiments. To train the models, we have tested 224 power devices from 10 different manufacturers. First, we describe a method to process the data for modeling purposes. Based on the in-house testing data, we implemented various ML models and observed that computational models such as Gradient Boosting and LSTM encoder-decoder networks can predict power device failure with high accuracy.

## Introduction

The most widely used power devices in the world utilize metal-oxide field effect transistor, or MOSFET technology [1]. Soon after their development in the early 70s, power MOSFETs were quickly adopted over the older bipolar junction transistors, as MOSFETs had faster switching speeds, lower switching power losses and reduced susceptibility to thermal runaway. Because of this new technology, power supplies can operate at higher frequencies and are increasingly smaller and lighter [2,3]. Power MOSFETs are found in almost any modern device, including consumer electronics, electric battery technology, mobile devices, data storage, radio, medical imaging, and more [4-9]. Power MOSFETs are also found in critical infrastructures like internet and telecommunications, automobiles, aerospace, and defense systems, making it difficult to overstate their importance in contemporary electronics.

Despite their ubiquity, power MOSFET devices are still susceptible to failure. Characterizing device reliability is vital to the successful design of an electronic system, and research in this field is vibrant. Investigators have studied reliability in several different formulations, such as failure in time (FIT), estimating device lifetime, and measuring a device's response to adversarial conditions [10-12]. However, even though research can inform system design in many different fields, power MOSFETs are often used in special applications, requiring domain specific research.

Space and aerospace are among the hardest domains for reliability research. In particular, exposure to ionizing radiation damages MOSFET devices in several ways, like radiation-induced gate leakage and drain current dispersion [13]. Moreover, space and aerospace environments often expose devices to extreme temperatures, making it necessary to determine a device's reliability when under thermal and radiological stress. Radiation testing is typically done with current-voltage ($I_g$ vs. $V_g$) curves. Experimenters collect these curves before irradiation and then re-collect them during different levels of exposure to look for changes in the characteristic [12]. These experiments require access to radiation sources and personnel, must be repeated for each device, and fail to gather information on thermal stressing. As a result, only a small fraction of the available devices are tested at a great expense.

Considering the difficulty of experimentally determining a device's response to thermal and radiological stress, we sought out a method for estimating stress response computationally. Recent results indicate that machine learning is capable of cutting-edge performance in the task of determining component reliability across multiple application domains, including electronics [14-18].

Submitted for review on July 16th, 2021. This work was supported in part by the Department of Energy (DoE), Office of Science, Office of Basic Energy Sciences, under DoE contract number DE-AC02-06CH11357
C. A. Olivares is with the Argonne National Laboratory, Lemont, IL 60439 USA (email: carlos.olivares.331@outlook.com)
R. Rahman is with the Argonne National Lab, Lemont, IL, 60439 (email: razeeebuet@gmail.com)
C. Stankus is with the Argonne National Lab, Lemont, IL 60439 USA (email: cstankus@anl.gov)
J. Hampton is with the Argonne National Lab, Lemont, IL 60439 USA (email: jayh1609@gmail.com)
A. Zedwick is with the Argonne National Lab, Lemont, IL 60439 USA (email: andzedwick@gmail.com)
M. Ahmed is with the Argonne National Lab, Lemont, IL 60439 USA (email: mahmed@anl.gov)

In fact, machine learning has already been successfully applied to MOSFET reliability in terms of failure classification and fault prognostics [19,20]. However, at the time of this writing, there is a relative lack of machine learning methods for establishing the radiation hardness of MOSFET devices.

In this paper, we describe machine learning approaches for estimating a MOSFET device's radiological and thermal reliability. We conduct experiments which collect data on both a device's static properties and its response to thermal and radiological stress. Leveraging data on its static properties, we formulate two different learning problems to predict a device's reliability and evaluate the predictive link between static properties and stress response. The efficacy and pitfalls of different machine learning methods are discussed in order to characterize the novel approach to radiation and thermal stress response, and our datasets are released publicly. We discuss our results and suggest opportunities to improve modeling.

## Data Collection and Objective

*Power Device Testing*

Any effort to estimate device reliability with machine learning suffers from a lack of a universal dataset, as MOSFET devices are manufactured by competing businesses. Moreover, experimental data on the radiation hardness of MOSFET devices is difficult to obtain, as they require access to radiation sources and personnel qualified to run radiation experiments.

In this project, we have tested devices from 10 different manufacturers. Experiments were conducted in the Los Alamos Neutron Science Center (LANSCE). Each device was tested in two stages.

*Dynamic Stress Experimentation*

Experiments were conducted with the 30L ICEHouse beam line at LANSCE. The procedure is as such: First, a power device would be turned off and a current measuring component would be placed in parallel with the transistor. Because the transistor device is turned off, all current is diverted to the measuring device, which records the current flowing through it. Then, the device was placed under one of two stress conditions: One stress condition is exposure to the neutron beam at 25ºC and the other is exposure to neutron radiation at 150ºC. If a device began to fail, it will allow current to flow through it, translating to less current flowing through the measuring device, which will be recorded. If a significant drop is seen in the current during irradiation, then the device has failed under stress. If not, then the device has passed.

*Pre and Post Stress Testing Measurements*

Before and after a device was placed in a stress condition, measurements were taken to observe the electrical and thermal behavior of the device. The pre-stress data characterizes a device's normal behavior, and the post-stress data can be used to detect changes in a device's behavior after being placed under thermal and radiological stress

*Experimental Data*

In supervised machine learning, we train models with labeled examples in data. With the data, a model can learn an underlying probability distribution which can be used to predict the labels of novel data points. In the case of our experimentation, the goal was to leverage the data collected during beam experimentation to predict the relibility of novel MOSFET devices. There are several formulations of this problem, and they are all discussed in subsection E.

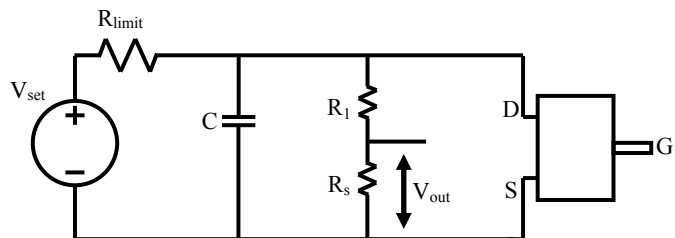

Fig. 1. A schematic of the circuit used during stress experimentation. During experimentation, the device is switched off, which diverts the current to the measuring device. If a device starts to fail, it will allow current to flow through it, and less charge will flow through the measuring device.

Beam experiments conducted at LANSCE yielded several different types of data, each serving a different purpose in each different formulation of the prediction problem.

*Static Data*

Pre-stress measurements demonstrates how gate to source voltage ($V_{gs}$) varies with gate to source current ($I_{gs}$) and vice-versa. Similarly, these measurements also show how drain to source voltage ($V_{ds}$) and current ($I_{ds}$) vary with respect to each other. Post-stress measurements collect the same data as pre-stress measurements, but *after* the dynamic stress experiment. This data will collect any noticeable changes in the tested device's static properties.

Pre- and post-stress static characterization measured three different parameters:
1. Threshold voltage, $V_{th}$ by measuring $V_{gs}$-$I_{gs}$ properties at a fixed $V_{ds}$.

2. $V_{gs}$-$I_{gs}$ properties at $V_{ds}$ = 0V.
3. $V_{ds}$-$I_{ds}$ properties at $V_{gs}$ = 0V.

*Stress Response (Dynamic) Data*

Stress response data was collected during beam testing and measures the behavior of an irradiated device in real time. Specifically, the data is a plot of charge ($I$ (A)) over time scaled units where time has been converted to neutron flux (n/cm$^2$). A significant drop in charge indicates that a device is unlikely to be reliable in the presence of significant ionizing radiation, and little to no reduction of current indicates a device is likely radiation hard.

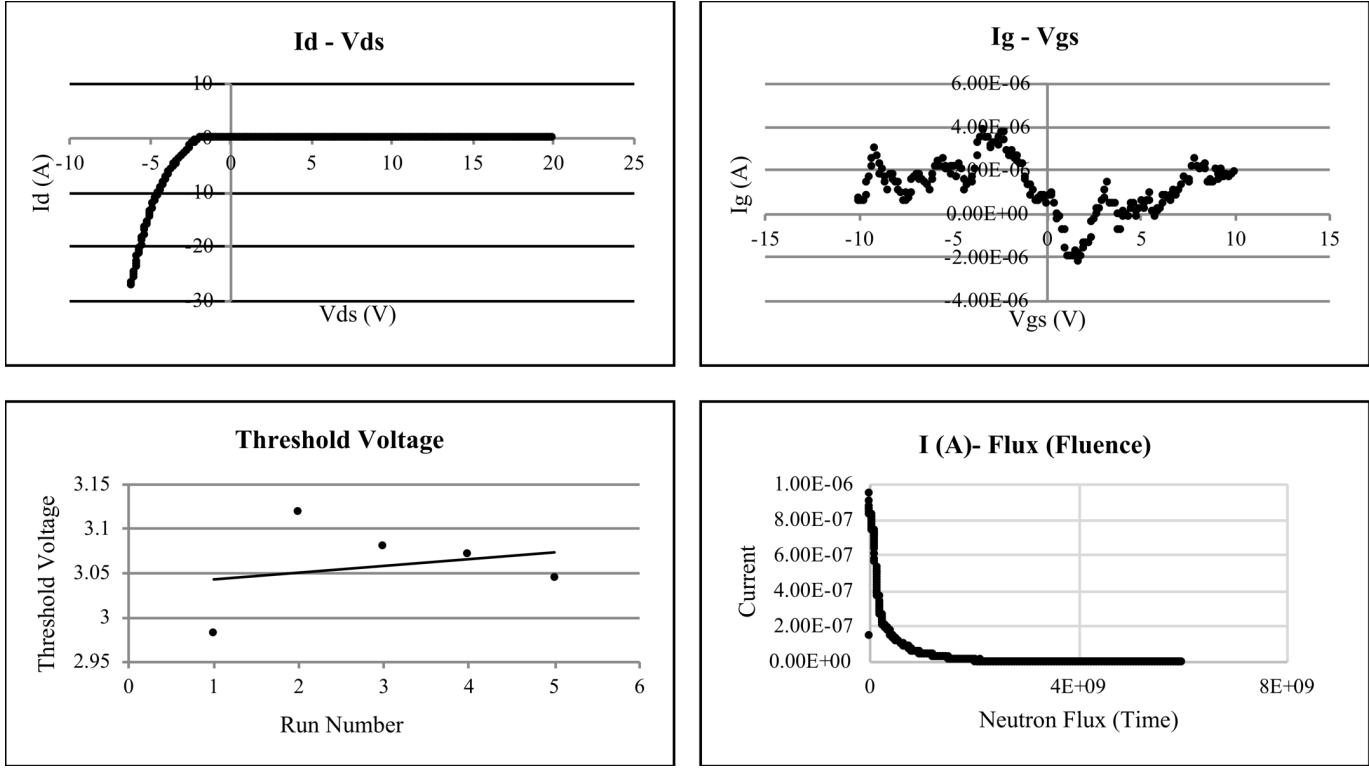

Fig. 2 The four different kinds of data collected during experimentation at LANSCE. Fig. 2(a) and 2(b) make up the static property data, which is leveraged to predict device stress response. Each point of each plot is used as an independent variable, and the dependent variables are points of the stress response curves like that of Fig. 2(d). Because each stress curve has thousands of points, equally spaced points were chosen as a subsample to recover the general trend of the curve. To gather threshold voltage data, we fixed drain to source voltage and measured $V_{gs}$-$I_{ds}$ properties.

*Failure in Time Data*

Failure in time (FIT) represents how many times a device fails every 10$^9$ hours. For example, a device labeled as 10-FIT is expected to fail 10 times for every 10$^9$ hours. This measure can be calculated from the experimental data. In particular, FIT can be calculated as

$$FIT = \frac{1}{MTBF} \times 10^9$$

where $MTBF$ is mean time between failure [21, Appendix B]. Since we collected time scaled data, we can recover $MTBF$. Accounting for dimensional analysis, the equation above can be re-written as

$$FIT = \frac{failed\ devices}{total\ devices * fluence} \times flux \times 60 \times 60 \times 10^9.$$

where $failed\ devices$ is the number of failed devices, $total\ devices$ is the total number of devices, $fluence$ is the final fluence value and $flux$ is amount of neutron flux.

*Outlier and Faulty Device Detection*

It is important to identify outlier devices, as using these devices in predictive modeling could jeopardize the reliability prediction. There are several categories of outlier devices identified which are shown with an example in Fig. 3 and listed below.

1. Little Data at Decline: Very few current points are provided for small amount of fluence which changed abruptly.
2. Current Jump: The stress response data has a discontinuous change in charge, resembling a jump discontinuity.

3. Instant Current Decline – Current has dropped drastically within low fluence values or time points.
4. Cloud: The stress response data has a significant number of data points with no relation to the data's general trend, resembling a function with several point discontinuities.
5. Odd Curve: The response data behaves in a way that defies any reasonable interpretation in the context of the experiment, such as an initial decrease in charge in, followed by a drastic increase in charge.
6. Odd Beginning Current: The charge measured at the beginning of the experiment is unexpected and has no justifiable explanation.

In addition to outlier devices, there are also outlier data points within each device's dynamic data. These points needed to be detected and filtered to avoid errors in any statistical calculations. An example of such outliers can be seen in the cloud of data points in Fig. 3(d). Among different outliers, some can be used for modeling while others compromise a model's accuracy if included during training.

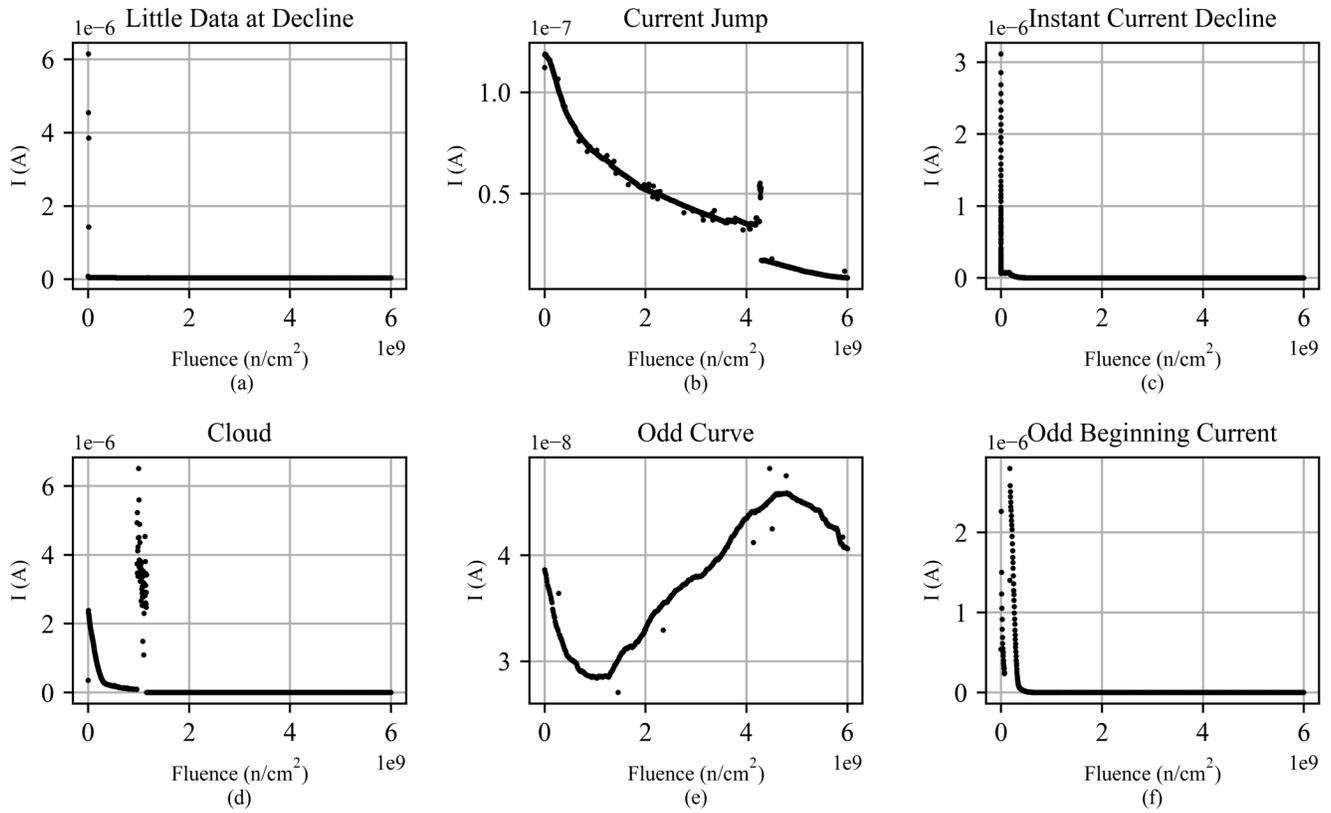

Fig. 3 There are six different kinds of outlier devices. Outlier devices are identified only through the behavior of the stress response curve and not through any of the static property data or threshold voltage data.

With outliers removed, it was important to set a standard fluence value for each device's data to stop at. After examining all the device's maximum fluence values, $6.0e9$ $n/cm^2$ was decided on. Devices that had data points recorded after that fluence value were cut short, with a fluence value added at exactly $6.0e9$ $n/cm^2$. Devices that did not have data points recorded up to that fluence had their data extended to that fluence. Any added data points were set at the same current value as the last current recorded before any points were added.

*Data Analysis*

After the experimental data was collected, we processed the data for the purposes of modeling. The most important part for generating model-applicable data is creating a benchmark x-axis data points. The benchmark points will indicate several factors: (I) the starting point of a device's static/dynamic data (II) ending point of a device's static/dynamic data (III) same x-axis points for all device data. By benchmarking these points, we can compare either static or dynamic data between devices. For example, after the benchmarking, currents will be calculated for same fluence/time points for all devices. To find the benchmark points, we selected starting and ending points of both static and dynamic data which can completely represent (end-to-end) the change in all devices' current. Along with that, we divided the middle points in a way which can accurately capture changes in the device data.

Before calculating benchmarked points, we performed interpolation on experimental data points, reducing noise in the data. We also performed extrapolation in cases where experimental data had not been calculated for benchmarked points. During the interpolation/extrapolation of experimental data, we fitted curves through the points. From these curves, the static or dynamic data of each device was calculated for benchmarked points.

Once the data was compiled into a pipeline file, minimal processing is needed for input into machine learning models. In fact, data processing for input to machine learning models consisted only of selecting appropriate portions of the pipeline file, and re-shaping the data into a tensor appropriate for training and inference. Some experiments also normalized numerical data, but none of the results presented in this paper originated from those experiments, and normalizing the data did not significantly alter the quality of results (accuracy of each algorithm's predictions).

*Problem Formulation*

In the broadest sense, our research aims to predict the probability of a device failure under stress conditions based on its electrical and thermal properties, and also provide an estimated range for the moment of failure. Different interpretations of device failure give rise to different formulations of the prediction problem, as do different combinations of data used to make predictions. One of the accepted measures of a device's failure is Failure-In-Time (FIT).

From our experiments, we can compute the FIT values by predicting two parameters: how many devices (per manufacturer) have failed, and what the(ir) final fluence value was. Both of these values are derived from fluence-current points, allowing for several different potential methods for establishing a prediction problem with machine learning:

- Case A: Predict success/failure of a device using static data and other variables such as temperature, bias voltage, etc. and subsequently calculate FIT.
- Case B: Predict current at different neutron flux/time points (stress response curve) using static data and other variables, determine success/failure of a device based on the predicted points/curve and subsequently calculate FIT.

**Direct Power Device Reliability Prediction**

Failure-in-time (FIT) is directly correlated with device pass-fail numbers. Due to that, we implemented models to predict whether a device failed or passed under a stress condition. In our initial analysis, we utilized static data ($V_{gs}$-$I_{gs}$, $V_{ds}$-$I_{ds}$), temperature and bias voltage to predict a device's pass-fail status.

| Device ID | Design Matrix | | | | | | Target Feature |
|---|---|---|---|---|---|---|---|
| | Manufacturer | Temperature | Bias Voltage | Average Threshold Voltage | $V_{gs} - I_{gs}$ | $V_{ds} - I_{ds}$ | Neutron Flux $-$ Current |
| $A_1$ | A | $25°C$ | 685 | 3.058 | 〰 | ⌐ | 〰 |
| $A_2$ | A | $150°C$ | 1027 | 1.354 | 〰 | ⌐ | — |
| $B_1$ | B | $25°C$ | 1369 | 2.069 | 〰 | ⌐ | — |
| $B_2$ | B | $150°C$ | 685 | 5.432 | 〰 | ⌐ | — |
| ⋮ | ⋮ | ⋮ | ⋮ | ⋮ | ⋮ | ⋮ | ⋮ |
| $K_N$ | K | $150°C$ | 1027 | 3.251 | 〰 | ⌐ | — |

Fig. 4 A diagram of the categorizations of different variables which have been employed as predictor and target for the machine learning models. The last two columns on the right area actually sequences showing how gate to source and drain to source charge and voltage change with respect to each other. The majority of the predictors data is actually sequence data of this kind, and prediction problem in formulation be can be thought of a sequence to sequence prediction problem. Naturally, this motivated us to approach this problem with recurrent neural networks, as they can recognize and work with sequence data. In fact, the encoder-decoder model we implemented predicted sequences as well. Despite this, recurrent neural networks were the weakest in terms of performance.

*Predicting Pass-Fail Status of Devices Using Traditional ML Algorithms*

For this classification problem we deployed multiple machine learning (ML) algorithms such as logistic regression, random forest and gradient boosting classification. Previously, these algorithms have shown success in binary classification, especially logistic regression which, has shown high classification accuracy for non-separable classes. Tree-based algorithms such as random forest or gradient boosting are efficient in classifying weak learners. Our data features a much higher proportion of failing devices than passing devices, and the pass-fail classes are inseparable. We chose algorithms which we believed were versatile enough to adapt to such data, and we also implemented several algorithms to determine which best captured the internal properties of the static properties data.

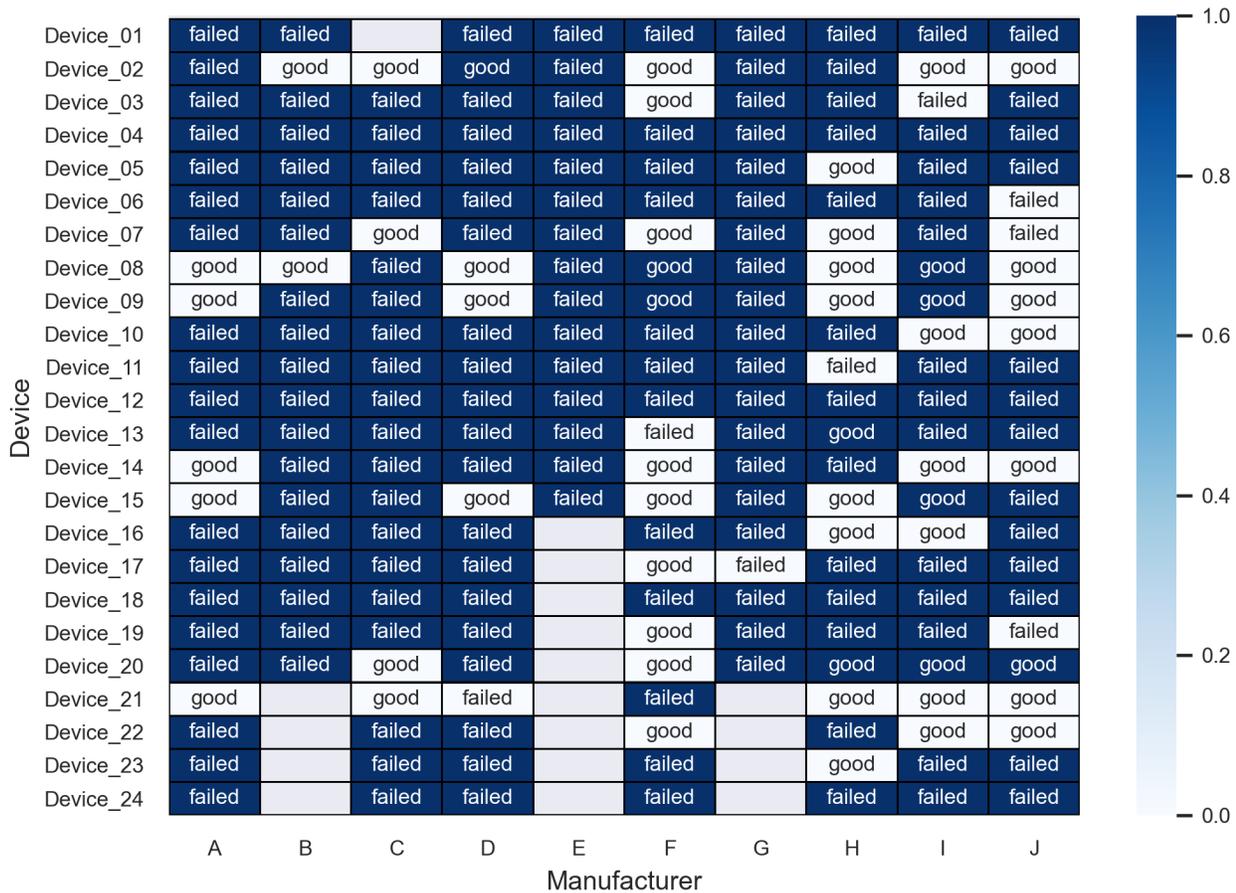

Fig. 5. Heatmap of Random Forest prediction results. Each box represents a device with a device number (row) and manufacturer (column), where the value in the box indicates the actual status of the device. If the prediction of that device is correct, the box is blue, otherwise white.

To validate models, we split the device data into 24-folds, stemming from the 24-devices being tested for each manufacturer apparatus (excluding outliers). For each fold $i$, device number $i$ from each manufacturer was used for test data, and all other devices for training data. Due to the variance of material properties of different manufacturers, we made device number the basis for folds instead of manufacturer. We used accuracy score for the classification metric , which indicates the proportion of the devices which were classified correctly from all devices.

Due to the imbalance of the data, our initial model results from logistic regression (average accuracy of 65%) skewed towards device failure, prompting us to implement balancing techniques which can even out the training data with oversampling (i.e., randomly duplicating passed devices), or undersampling (i.e., randomly deleting failed devices), or SMOTE (Synthetic Minority Sampling Technique; i.e., a subcategory of oversampling in which data for passed devices are synthetically generated) [22]. Unfortunately, none of these data balancing methods significantly improved logistic performance. Compared to logistic regression, tree-based algorithms random forest and gradient boosting classification have performed well with classification accuracy of 76.1% and 75.1% respectively.

Fig. 5 shows the RF classification results of each devices are shown, from which we have observe that models are highly unsuccessful in classifying passing devices even with class balancing techniques. Fig. 6 shows confidence intervals for classification accuracy of different algorithms, which indicates that performance varies significantly with different manufacturers, even though data of each class are included in the training.

*Predicting Pass-Fail Status of Devices Using Multi-Step ML Algorithms*

Our results so far indicated to us that directly predicting pass-fail status was unreliable, due to the variability of device properties with respect to manufacturer. This behavior became evident when we characterized the pre-stress device data using principal component analysis and hierarchical clustering (Fig. 7). Furthermore, device data are clustered based on device properties such as voltage rating and device types. By optimizing this behavior, we have built a multi-step prediction model, where

1. In the first step of the algorithm, we have predicted the properties of the testing device using all available training devices.
2. In the second step, we have predicted the reliability of the device by only using training devices representing testing device properties.

With this algorithm, it will be possible to capture the internal properties of the device to get better predictions, along with forecasting the service-life of a device in an extreme environment with better accuracy.

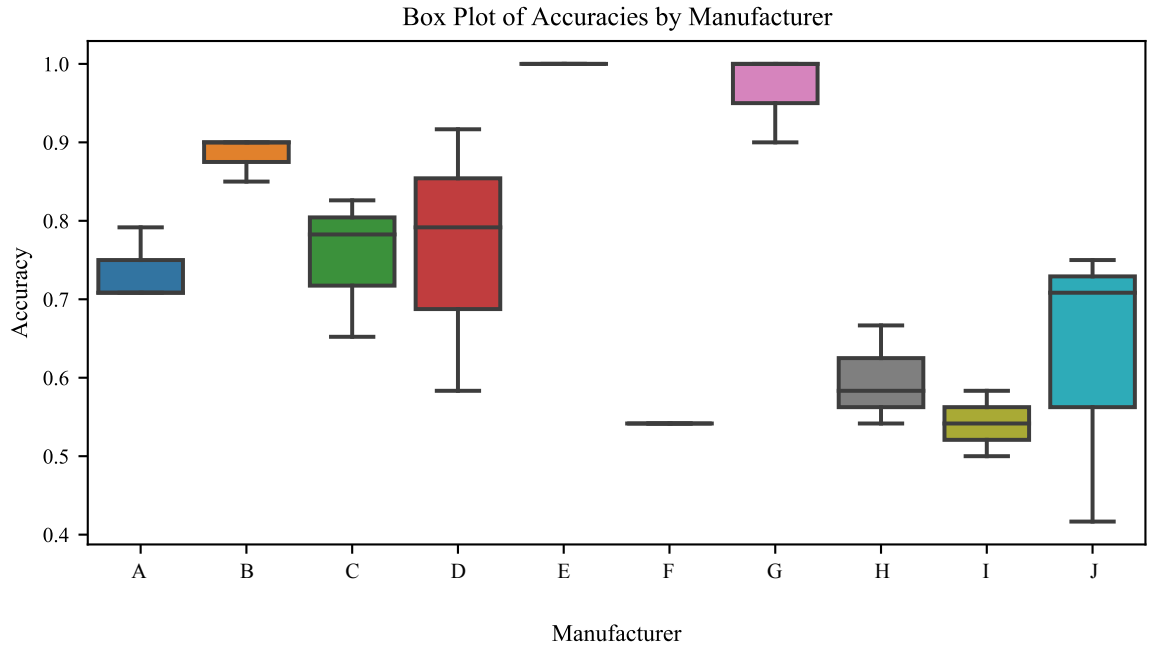

Fig. 6. Device pass-fail accuracies predicted by different machine learning algorithms (logistic regression, random forest classification, and gradient boosting classification). Box plot is representing the range of accuracies by different algorithms.

Similar to the single step reliability prediction approach, devices are split using 24-fold cross validation method. During the first-step of this algorithm, we have predicted manufacturers/voltages of testing device, and with the exception of two manufacturers (I and J; Fig. 7(a)), prediction of manufacturers/voltages were accurate. These successful results indicate that data from each manufacturer is distinctive and if we need to predict a new manufacturer's device reliability, we need to consider the batch effect.

In the second-step of this algorithm, device failure is predicted only on devices whose manufacturer or voltage was correctly predicted. In this step, the model was trained on devices which belong to the first step manufacturer/voltage predicted class, and only one device was used for testing at a time. For example, if in the first-step of prediction, manufacturers/voltages are predicted as manufacturer A, then only manufacturer A devices (except a single testing device) will be used for training the model to predict the testing device reliability. On average, when manufacturers or voltages were predicted first, the accuracy of device failure predictions was 69.1% when the prediction model was Random Forest. The accuracy is lower compared to direct reliability prediction approach due to the decreased amount of available data in the training set.

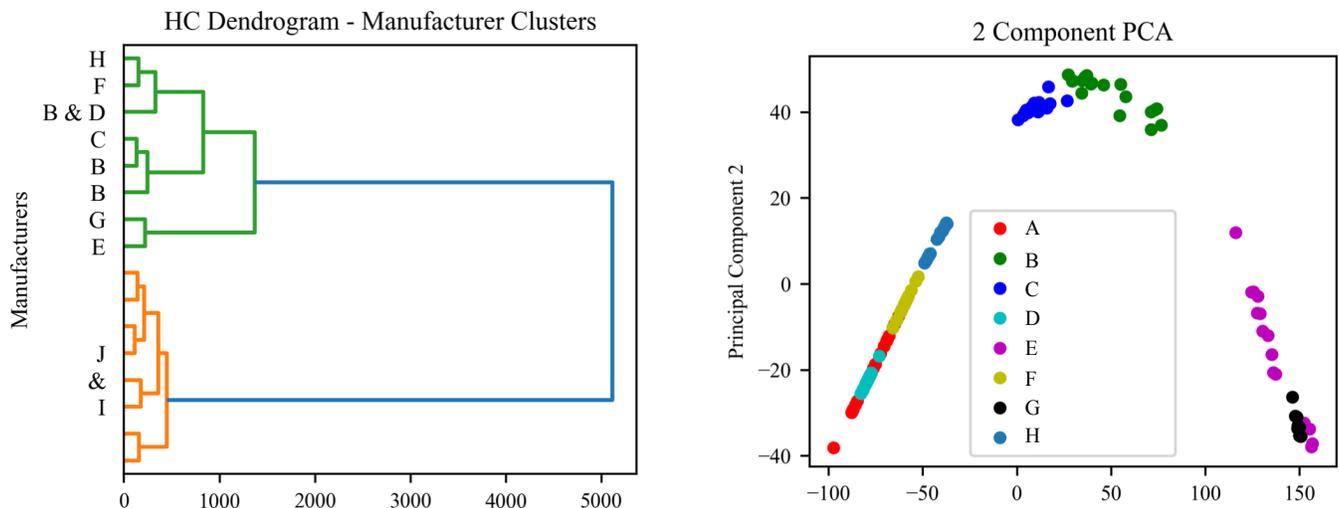

Fig. 7. Results from a hierarchical and a two component PCA of the devices. Our algorithms predicted poorly for manufacturers I and J, so they were removed for two component PCA.

## Indirect Power Device Reliability Prediction

Instead of directly predicting power device reliability (pass-fail status), models can instead predict current values at different neutron fluence points using static data. With these points, we can determine reliability of devices or calculate FIT values. This methodology allows us to not only determine the status of device, but also can observe the changes of currents of devices with respect to time.

*Stress Response Prediction using Gradient Boosting Algorithm*

Unlike the previous reliability models in which targets are binary (pass-fail status), stress response data consists of multiple fluence/time points, where currents in each of these points for different devices are discrete (Fig. 2). So, to predict the stress response curve, we need to train multiple models where each model will be trained and tested on a specific fluence point. By concatenating the predicted currents along the sampled fluence points (in chronological order), we can recover the stress response curve.

Previously, tree-based algorithms such as gradient-boosting algorithm have shown success in predicting continuous values [23,24]. The gradient boosting algorithm's accuracy comes from constructing a strong learner from an ensemble of 'weak' learners. Let $y_{f,d}$ be a measured current for some neutron flux point $f$ and device $d$. Define a loss function $L(y_{f,d}, \hat{y})$ which evaluates how similar the measured value and the predicted value $\hat{y}$ are. In this prediction process, the goal is to produce a hypothesis function $H$ of the device's static property data $x_d$ which minimizes the loss function. Gradient boosting constructs a learner in $M$ stages. Each stage begins with a weak learner $H_i$ for $1 \leq i \leq M$, to which the gradient boosting algorithm adds a new $h_i$, re-writing the learning problem as

$$H_{i+1}(x_d) = H_i(x_d) + h_i(x_d) = y_{f,d}.$$

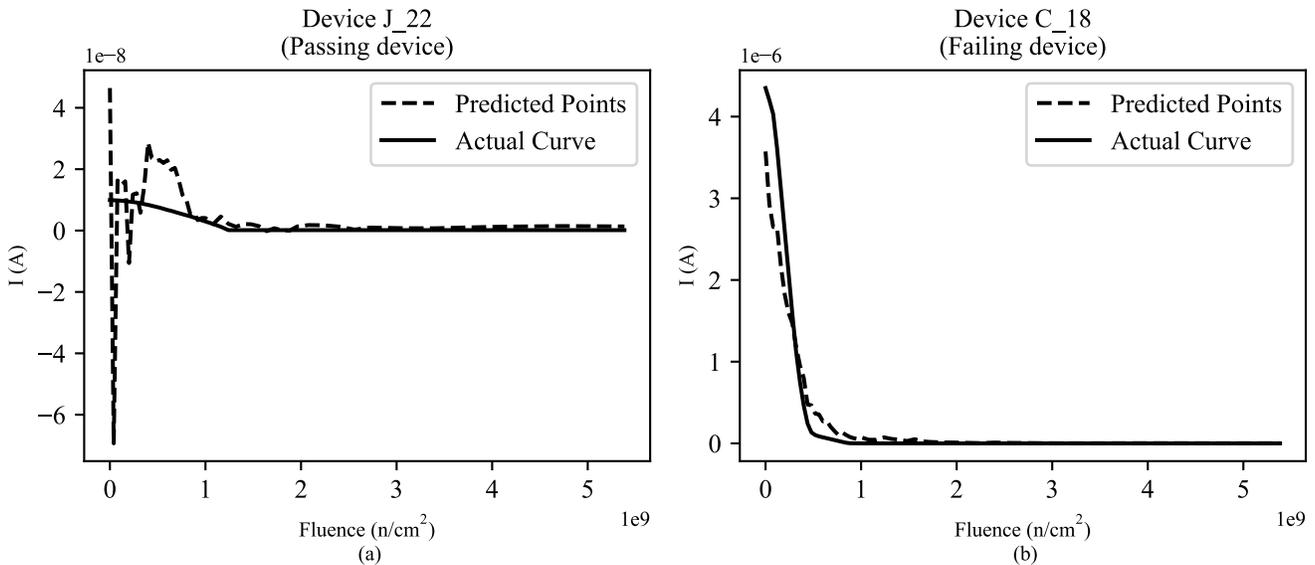

Fig. 8. In each plot, sampled points of a device's stress response curve are shown along with their corresponding prediction from the gradient boosting models. The models predict the general trajectory of the measured response curve accurately in the majority of cases.

More importantly, the new estimator can be defined by $h_i(x_d) = y_{f,d} - H_i(x_d)$, which shows that gradient boosting fits to the residual difference between the prediction from $H_i$ and the observed value. As a result of repeatedly seeking to reduce the residual difference, gradient boosting models excel at predicting current at a specific neutron flux point.

Through setting up gradient boosting models to predict current at several neutron flux values, it was possible to predict the general trajectory of a device's stress response curve. Predicting the stress response in this way provided the benefit of having several methods to evaluate the accuracy of predictions.

*Evaluating the Predicted Stress Response Curves*

For this project, two different evaluation methodologies were developed. One method uses the predicted curve to compute a pass-fail status and compare it to the observed pass-fail status, and the other is a pointwise evaluation to determine if the predicted curve is an acceptable approximation to the measured stress response curve. Consequently, the results from one kind of evaluation system cannot be directly compared to those from a different kind. In order to be best informed regarding model performance, both kinds of grading systems were developed, including a strict grading system to evaluate the curve as an approximation to the measured curve, a relaxed grading system which computed and compared pass or failed statuses, and a detailed algorithm to determine if a curve is reasonably close to the measured curve. We also implemented an algorithm that treated the predicted points as measured stress response curves and computed several values from them, such as FIT and pass-fail status. We call this method generalized grading. In Fig. 8, we show two situations (an actual passing and failing device) where our modeling approach prediction was able to capture the actual stress response (fluence-current) curve characteristics. In general, using the (xtreme) gradient boosting (XGB) approach, we have got an accuracy of ~80% of predicting power device reliability (pass-fail status) where passing or failing was determined based on the predicted dynamic data curve. In supplementary Fig. 5, prediction results from each device are presented by a heatmap.

*Stress Response Prediction with Recurrent Neural Networks*

Although gradient boosting provided strong results, there were several aspects of the static properties data that gradient boosting models did not make use of. A device's static property data is actually comprised of sequences which describe how charge changes with respect to voltage and vice versa, but the gradient boosting models can not recognize the data as sequence data, and treats each timestep as an individual feature. Moreover, gradient boosting models can only predict current at a single point of neutron flux, when the target data is actually a stress response curve.

Recurrent neural networks overcome the limitations of gradient boosting models. By utilizing an LSTM based encoder-decoder model, the prediction problem was able to recognize the sequential structure of the predictor data, and predict the entire stress response curve at once. The encoder-decoder architecture is designed for sequence to sequence translation involving two components: an encoder which embeds an input sequence into a fixed length vector, and a decoder which uses the embedding to output a predicted sequence.

Before approaching the problem with the encoder decoder models, all previous deep learning models which we utilized to predict stress curves failed to learn. However, with the implementation of a Long Short-Term Memory (LSTM) based encoder-decoder model, the predicted curves indicated that the model was learning the data sequence, despite being far from the measured stress experiment curve (supplementary Fig. 6).

## Discussion

The main objective of this paper is to create a generalized algorithm for power device reliability prediction. Currently, power devices are generated without a precise idea of the service-life at harsh conditions. Therefore, application engineers need to conduct time-consuming testing for a particular mission specific operation. With our algorithm, the need for experimentally validating devices is reduced considerably. This approach will not only reduce the cost of reliability testing, but also hazard of radiation from it.

In general, machine learning relies heavily on the amount of data. There are no public databases for power device reliability in stress conditions. In this situation, we conducted our own power device testing for thermal and radiological stress conditions. However, conducting stress tests is very time consuming and requires access to sophisticated radiology labs such as LANSCE. We were allowed access to LANSCE for a very limited amount of time, and were not able to collect sufficient data to make the most of machine learning methods. Our experiments only measured the response of 240 devices, translating to only 224 data points for training and testing models once we accounted for outliers. Many traditional machine learning algorithms do not reach their full predictive capacity until they experience more than double the number of data points. All deep learning methods in particular usually require thousands to tens of thousands of data points to maximize their predictive capability.

Models which predicted the stress responses curve have the original goal of determining power device reliability. But, from the predicted stress response curve, researchers can determine the time point (fluence point) in which the device begins to fail (when current drops). This information can assist an application engineer during experiment design, and stop dangerous device failure accidents.

Predictions were made on the basis of leveraging data collected on the static properties of a device to predict its response to thermal and radiological stress, which is measured as a dynamic property. Ultimately, this link was more useful to some models than others. Deep learning methods in particular were not able to utilize static property data for accurate predictions to any appreciable degree. Gradient boosting models, however, were able to gain some predictive ability with the data.

During beam experimentation, a device was labeled as failing or passing depending on the measure of voltage dropped below a threshold value. However, choosing a threshold value is subjective to a certain degree. As a result, the stress response curves of many passing devices were very similar to those of failing curves. Because of a lack of a clear difference between stress curves of passing and failing devices, curve predicting models often were less sensitive to passing devices. Moreover, the lack of a clear distinction between passing and failing introduces ambiguity to the eventual goal of binary classification of passing or failing, whether it be through curve evaluating or from directly predicting with models.

10The results show that predicting the stress response (dynamic) data is more effective than predicting a device's pass-fail status directly. However, the most effective method developed was not by predicting the actual stress response curve, but by through several different prediction problems to predict different points of the stress response curve. We desire to develop a method that could predict the stress response curve with just one learning problem and model. The current encoder-decoder method, while not the strongest in performance, does show promise, and can possibly be improved in future work.

## Conclusion

In this paper, we introduced a novel application of machine learning to power device reliability prediction. Using data collected on the static properties of a MOSFET device, our models predicted its response to extreme temperatures and ionizing radiation. Results from our predictions evaluate the capacity of a device's static properties to serve as a predictor for stress response. Although this predictive capacity seems insufficient for directly predicting if a device passes or fails a stress test, we show that static properties form a robust predictive link to measured current at specific neutron flux points during stress testing, allowing researchers to recover a device's stress response. The gradient boosting algorithm best approximated true stress response, and sequence to sequence prediction efforts such as the encoder-decoder model can further improve the prediction. We have made all the experiment data along with generalized algorithm (as command-line-interface) publicly available (TBA) which can be used to further improve the prediction result or predict power device reliability without extensive knowledge of the algorithms.

## Acknowledgement

The work is supported by U. S. Department of Energy (DoE), Office of Science, Office of Basic Energy Sciences, under DoE contract number DE-AC02-06CH11357. This work was supported in part by the U.S. Department of Energy, Office of Science, Office of Workforce Development for Teachers and Scientists (WDTS) under the Science Undergraduate Laboratory Internships Program (SULI).## References

[1] *Power MOSFET Basics*, Alpha & Omega Semiconductor, Sunnyvale, CA.
[2] *Power Supply Technology - Buck DC/DC Converters*, Mouser Electronics, Mansfield, TX.
[3] D. A. Grant and J. Gowar, *Power MOSFETs: Theory and Applications*, New York, NY, USA: J. Wiley, 1989.
[4] *Semiconductor solutions for healthcare applications*, ST Microelectronics, Quakertown, PA, 2019.
[5] F. Maloberti and A. C. Davies, "Switched Capacitor Filters," in *A Short History of Circuits and Systems*. Aalborg: River Publishers, 2016. [Online]. Available: http://www.vlebooks.com/vleweb/product/openreader?id=none&isbn=9788793379695
[6] J.-P. Colinge and J. Greer, *Nanowire transistors: physics of devices and materials in one dimension*. Cambridge, United Kingdom: Cambridge University Press, 2016.
[7] S. Z. Asif, *5G Mobile Communications Concepts and Technologies*, 1st ed., Boca Raton, FL: CRC Press/Taylor & Francis: CRC Press, 2018.
[8] H. J.M. Veendrick, *Nanometer CMOS ICs: From Basics to ASICs*, Cham: Springer International Publishing, 2017.
[9] S. W. Amos and M. James, *Principles of transistor circuits: introduction to the design of amplifiers, receivers, and digital circuits*, 9th ed., Woburn, MA: Newnes 2000.
[10] Z. Ni, X. Lyu, O. P. Yadav, and D. Cao, "Review of SiC MOSFET based three-phase inverter lifetime prediction," in *2017 IEEE Applied Power Electronics Conference and Exposition (APEC)*, Tampa, FL, USA, Mar. 2017, pp. 1007–1014. doi: 10.1109/APEC.2017.7930819.
[11] D. R. Ball et al., "Estimating Terrestrial Neutron-Induced SEB Cross Sections and FIT Rates for High-Voltage SiC Power MOSFETs," *IEEE Trans. Nucl. Sci.*, vol. 66, no. 1, pp. 337–343, Jan. 2019, doi: 10.1109/TNS.2018.2885734.
[12] M. G. Faruk et al., "Proton and neutron radiation effects studies of MOSFET transistors for potential deep-space mission applications," in *2012 IEEE Aerospace Conference*, Mar. 2012, pp. 1–13. DOI: 10.1109/AERO.2012.6187016.
[13] M. H. Wong et al., "Radiation hardness of β -Ga 2 O 3 metal-oxide-semiconductor field-effect transistors against gamma-ray irradiation," *Appl. Phys. Lett.*, vol. 112, no. 2, p. 023503, Jan. 2018, doi: 10.1063/1.5017810.
[14] A. Alghassi, S. Perinpanayagam, and M. Samie, "Stochastic RUL Calculation Enhanced With TDNN-Based IGBT Failure Modeling," *IEEE Transactions on Reliability*, vol. 65, no. 2, pp. 558–573, Jun. 2016, DOI: 10.1109/TR.2015.2499960.
[15] E. F. Alsina, M. Chica, K. Trawiński, and A. Regattieri, "On the use of machine learning methods to predict component reliability from data-driven industrial case studies," Int J Adv Manuf Technol, vol. 94, no. 5, pp. 2419–2433, Feb. 2018, DOI: 10.1007/s00170-017-1039-x.
[16] Y. Zheng, L. Wu, X. Li, and C. Yin, "A relevance vector machine-based approach for remaining useful life prediction of power MOSFETs," in *2014 Prognostics and System Health Management Conference (PHM-2014 Hunan)*, Zhangjiaijie City, China, Aug. 2014, pp. 642–646. doi: 10.1109/PHM.2014.6988252.
[17] X. Fang, S. Lin, X. Huang, F. Lin, Z. Yang, and S. Igarashi, "A review of data-driven prognostic for IGBT remaining useful life," *Chinese Journal of Electrical Engineering*, vol. 4, no. 3, pp. 73–79, Sep. 2018, DOI: 10.23919/CJEE.2018.8471292.
[18] W. Chen, L. Zhang, K. Pattipati, A. M. Bazzi, S. Joshi, and E. M. Dede, "Data-Driven Approach for Fault Prognosis of SiC MOSFETs," *IEEE Transactions on Power Electronics*, vol. 35, no. 4, pp. 4048–4062, Apr. 2020, DOI: 10.1109/TPEL.2019.2936850.
[19] D. McMenemy, W. Chen, L. Zhang, K. Pattipati, A. M. Bazzi, and S. Joshi, "A Machine Learning Approach for Adaptive Classification of Power MOSFET Failures," in 2019 IEEE Transportation Electrification Conference and Expo (ITEC), Jun. 2019, pp. 1–8. doi: 10.1109/ITEC.2019.8790564.
[20] V. Ferlet-Cavrois et al., "Statistical Analysis of Heavy-Ion Induced Gate Rupture in Power MOSFETs—Methodology for Radiation Hardness Assurance," IEEE Transactions on Nuclear Science, vol. 59, no. 6, pp. 2920–2929, Dec. 2012, DOI: 10.1109/TNS.2012.2223761.
[21] M. A. Levin and T. T. Kalal, *Improving Product Reliability: Strategies and Implementation*. Chichester, UK: John Wiley & Sons, Ltd, 2003. doi: 10.1002/0470014024.
[22] N. V. Chawla, K. W. Bowyer, L. O. Hall, and W. P. Kegelmeyer, "SMOTE: Synthetic Minority Over-sampling Technique," *jair*, vol. 16, pp. 321–357, Jun. 2002, doi: 10.1613/jair.953.
[23] J. H. Friedman, "Stochastic gradient boosting," *Computational Statistics & Data Analysis*, vol. 38, no. 4, pp. 367–378, Feb. 2002, doi: 10.1016/S0167-9473(01)00065-2.

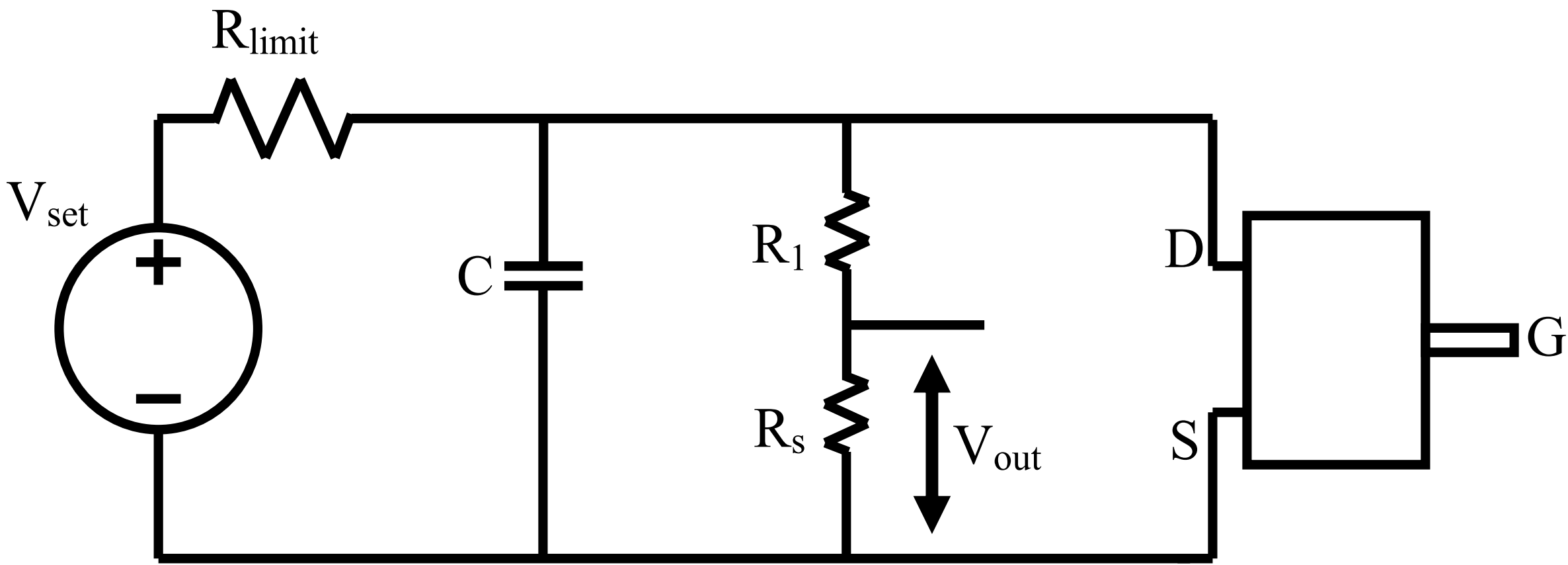

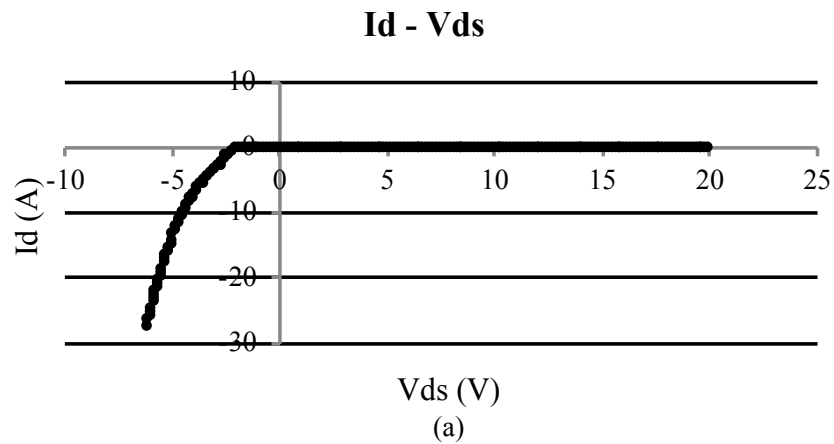
(a)

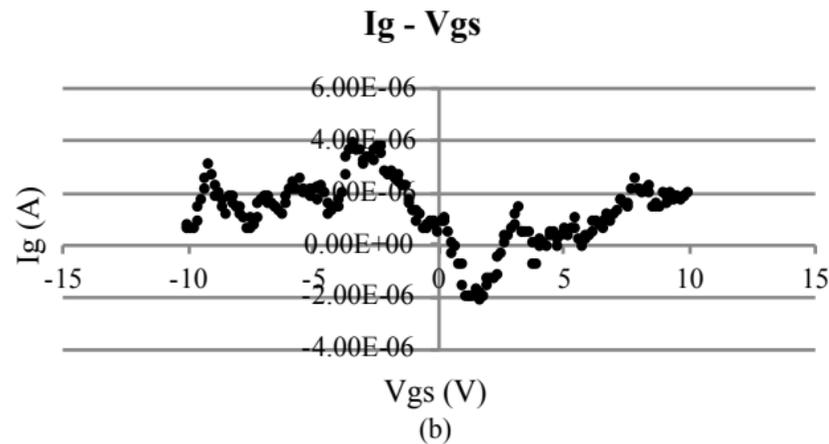
(b)

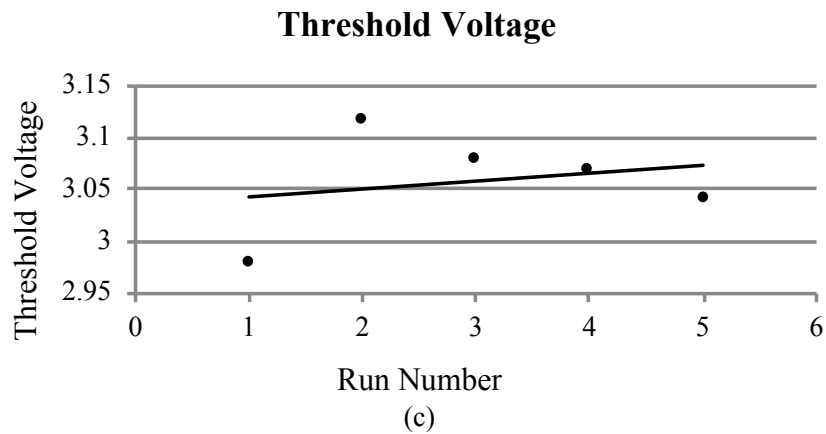
(c)

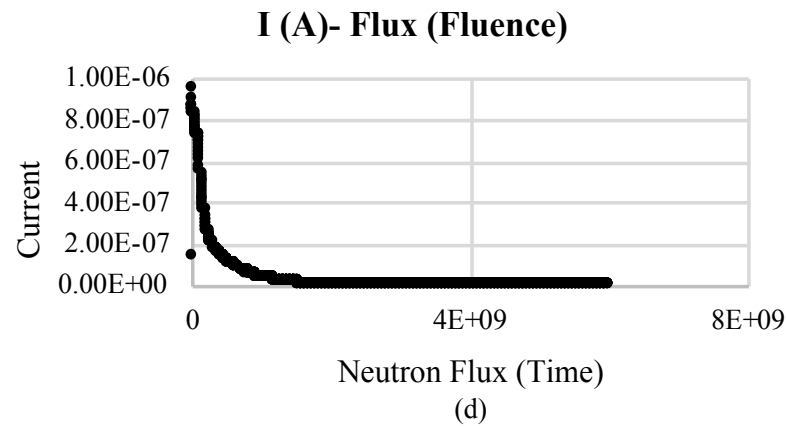
(d)

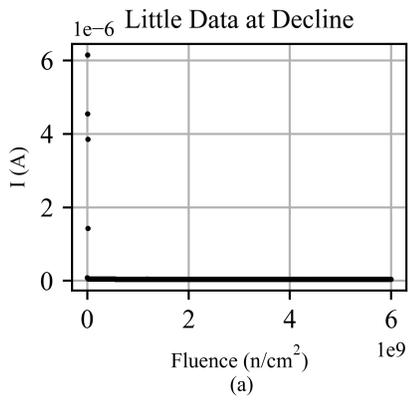
(a) Little Data at Decline

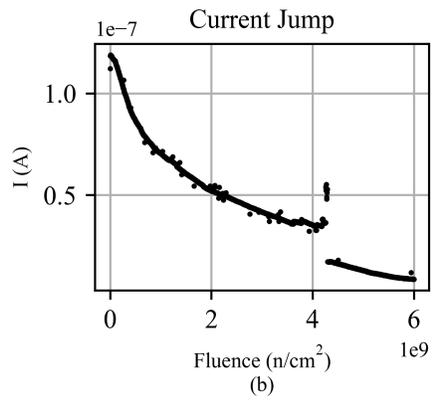
(b) Current Jump

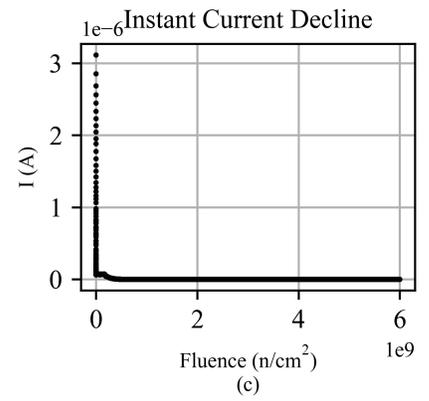
(c) Instant Current Decline

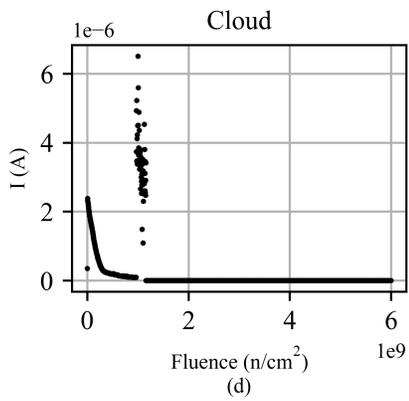
(d) Cloud

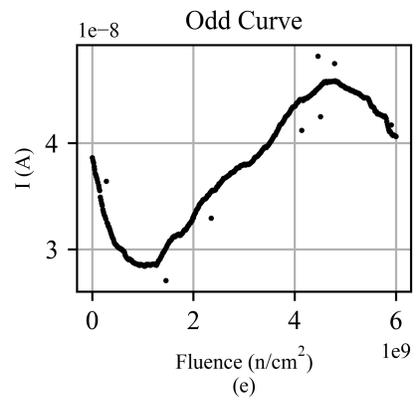
(e) Odd Curve

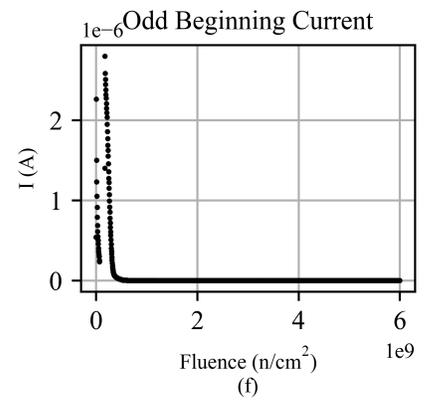
(f) Odd Beginning Current

| Device ID | Design Matrix | | | | | | Target Feature |
| --- | --- | --- | --- | --- | --- | --- | --- |
| | Manufacturer | Temperature | Bias Voltage | Average Threshold Voltage | $V_{gs} - I_{gs}$ | $V_{ds} - I_{ds}$ | Neutron Flux $- Current$ |
| $A_1$ | A | $25°C$ | 685 | 3.058 | 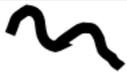 | 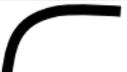 | 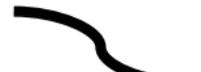 |
| $A_2$ | A | $150°C$ | 1027 | 1.354 | 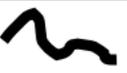 | 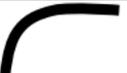 | 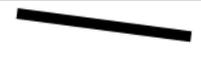 |
| $B_1$ | B | $25°C$ | 1369 | 2.069 | 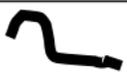 | 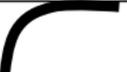 | 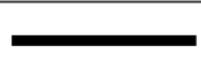 |
| $B_2$ | B | $150°C$ | 685 | 5.432 | 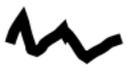 | 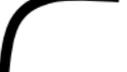 | 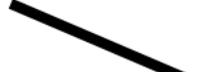 |
| ⋮ | ⋮ | ⋮ | ⋮ | ⋮ | ⋮ | ⋮ | ⋮ |
| $K_N$ | K | $150°C$ | 1027 | 3.251 | 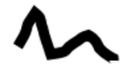 | 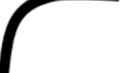 | 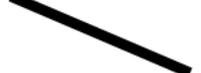 |

| Device | A | B | C | D | E | F | G | H | I | J |
|---|---|---|---|---|---|---|---|---|---|---|
| Device_01 | failed | failed |  | failed | failed | failed | failed | failed | failed | failed |
| Device_02 | failed | good | good | good | failed | good | failed | failed | good | good |
| Device_03 | failed | failed | failed | failed | failed | good | failed | failed | failed | failed |
| Device_04 | failed | failed | failed | failed | failed | failed | failed | failed | failed | failed |
| Device_05 | failed | failed | failed | failed | failed | failed | failed | good | failed | failed |
| Device_06 | failed | failed | failed | failed | failed | failed | failed | failed | failed | failed |
| Device_07 | failed | failed | good | failed | failed | good | failed | good | failed | failed |
| Device_08 | good | good | failed | good | failed | good | failed | good | good | good |
| Device_09 | good | failed | failed | good | failed | good | failed | good | good | good |
| Device_10 | failed | failed | failed | failed | failed | failed | failed | failed | good | good |
| Device_11 | failed | failed | failed | failed | failed | failed | failed | failed | failed | failed |
| Device_12 | failed | failed | failed | failed | failed | failed | failed | failed | failed | failed |
| Device_13 | failed | failed | failed | failed | failed | failed | failed | good | failed | failed |
| Device_14 | good | failed | failed | failed | failed | good | failed | failed | good | good |
| Device_15 | good | failed | failed | good | failed | good | failed | good | good | failed |
| Device_16 | failed | failed | failed | failed |  | failed | failed | good | good | failed |
| Device_17 | failed | failed | failed | failed |  | good | failed | failed | failed | failed |
| Device_18 | failed | failed | failed | failed |  | failed | failed | failed | failed | failed |
| Device_19 | failed | failed | failed | failed |  | good | failed | failed | failed | failed |
| Device_20 | failed | failed | good | failed |  | good | failed | good | good | good |
| Device_21 | good |  | good | failed |  | failed |  | good | good | good |
| Device_22 | failed |  | failed | failed |  | good |  | failed | good | good |
| Device_23 | failed |  | failed | failed |  | failed |  | good | failed | failed |
| Device_24 | failed |  | failed | failed |  | failed |  | failed | failed | failed |

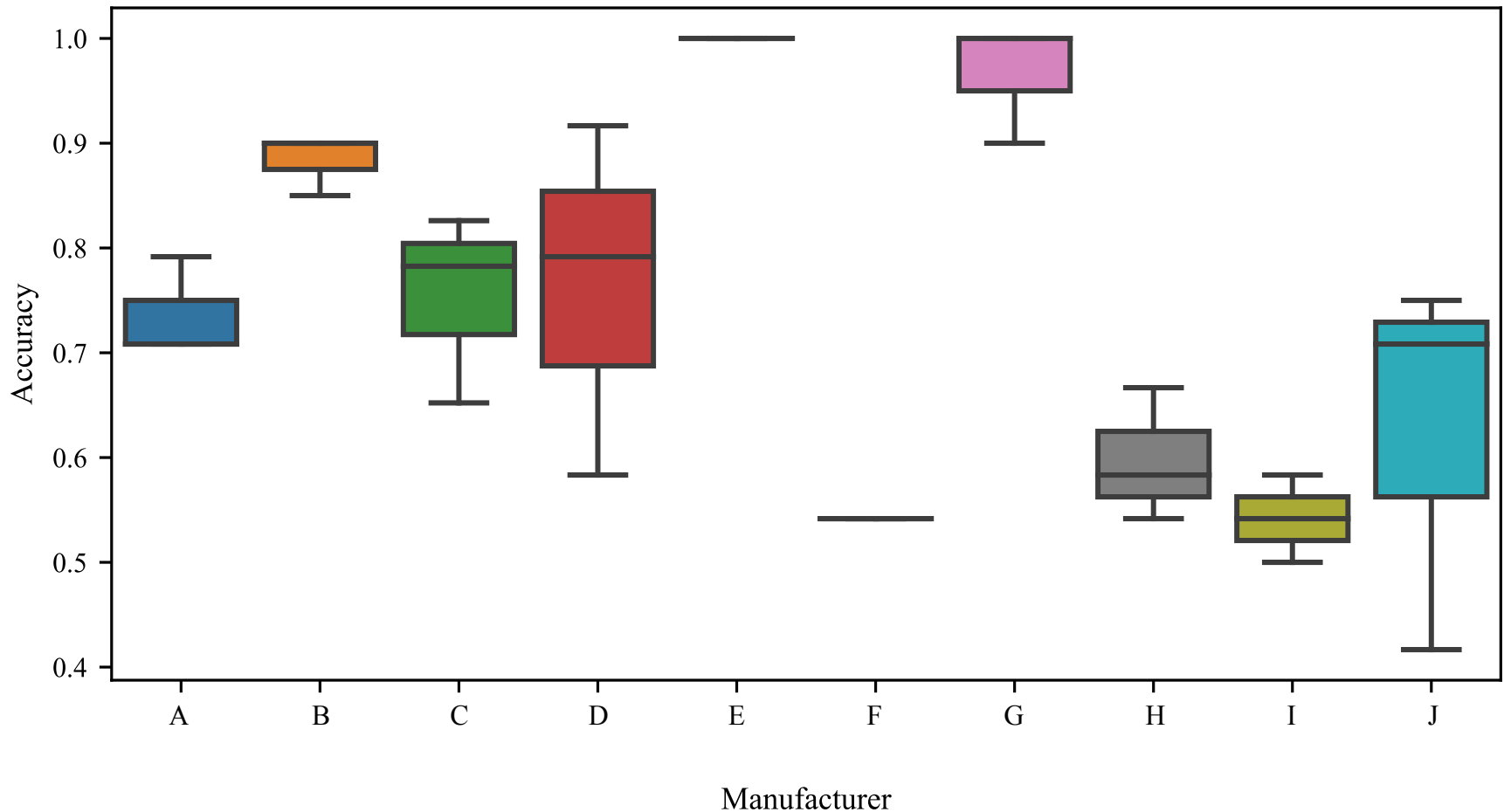

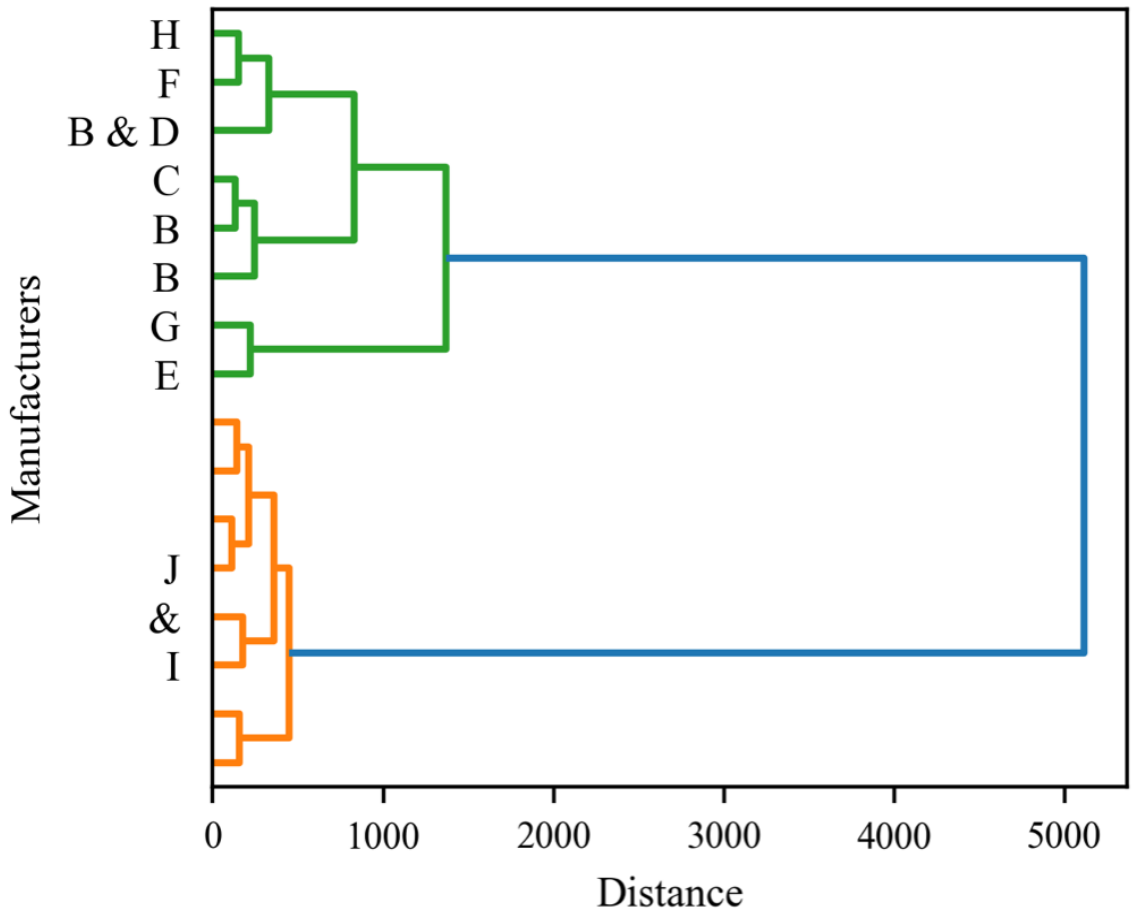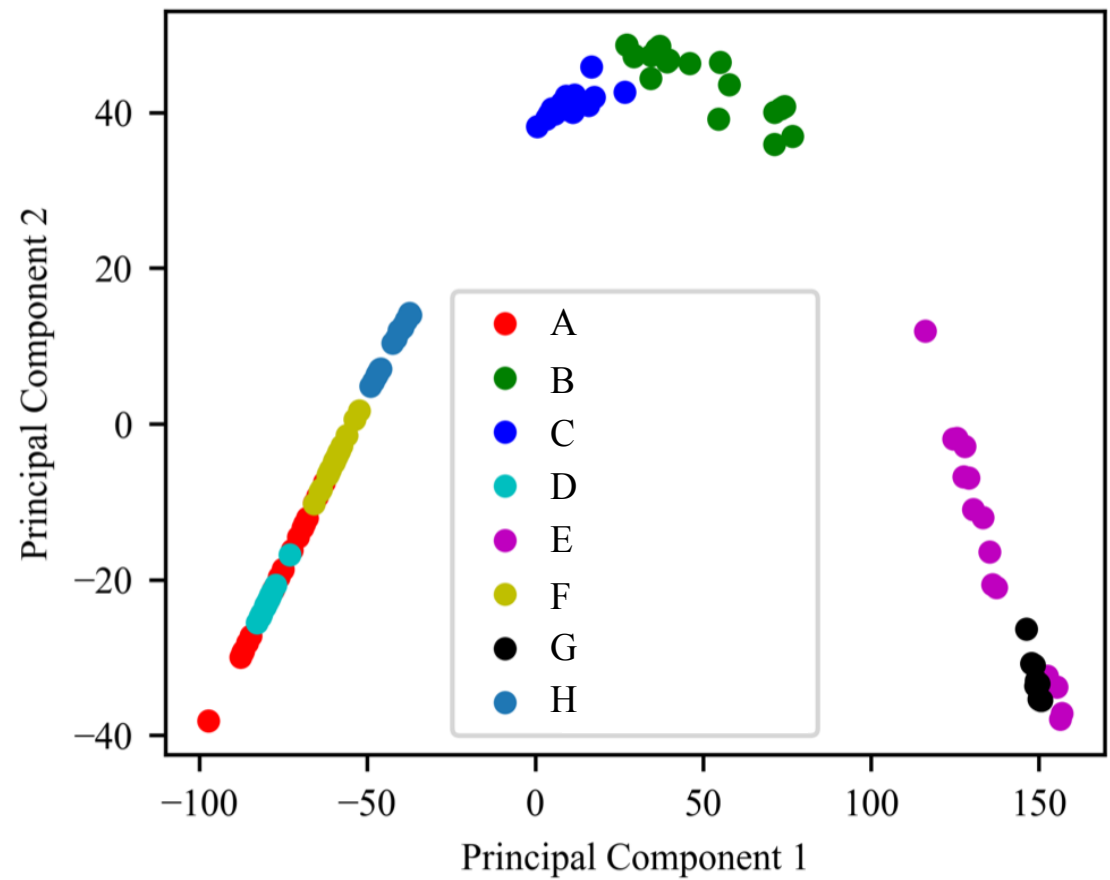

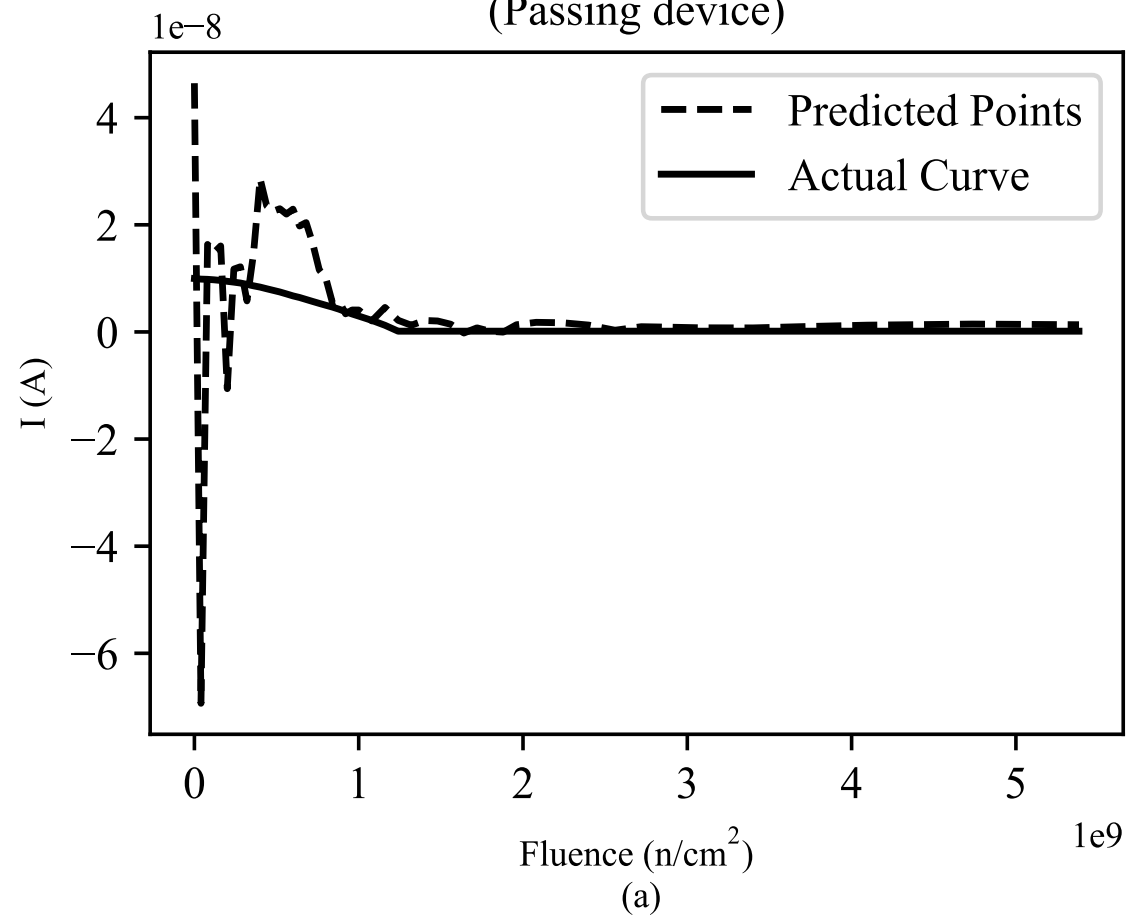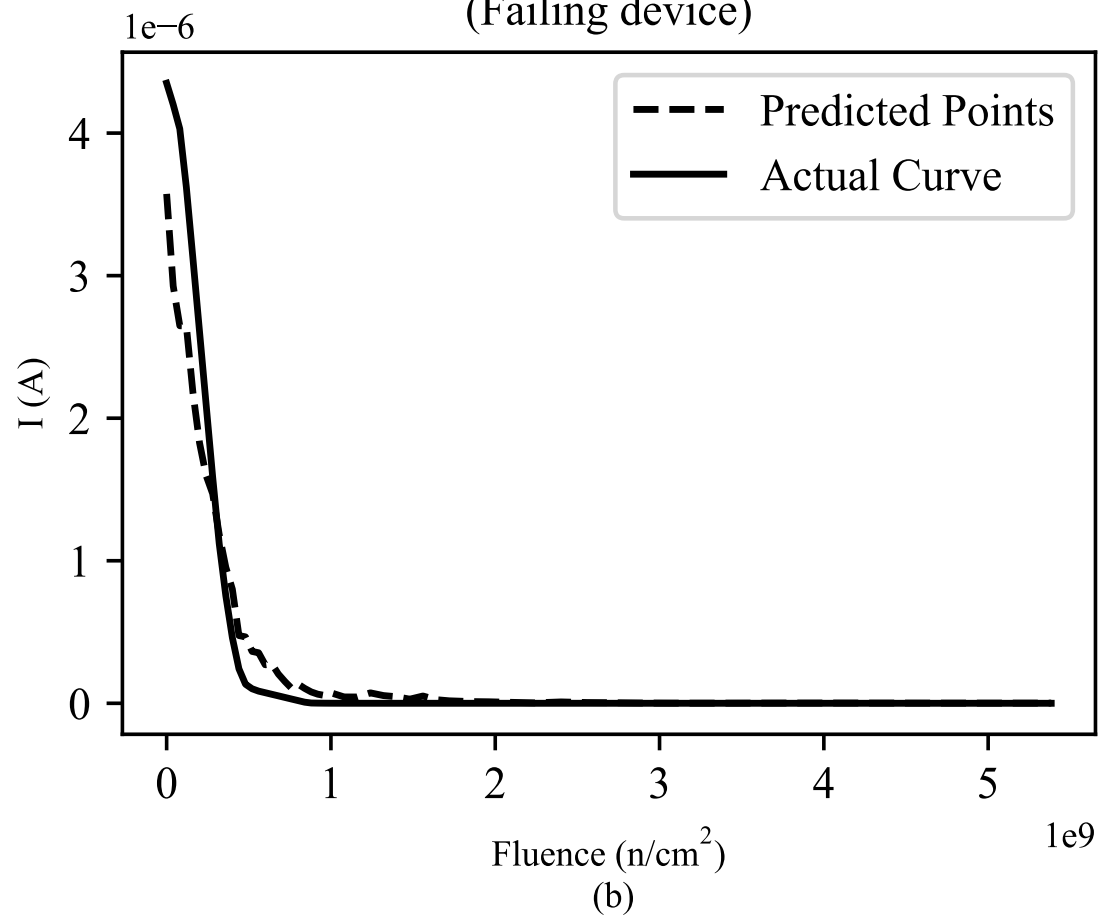